\input{access-engine-compat.tex}
\documentclass[nolineno]{ieeeaccess}
\usepackage[nocompress]{cite}
\usepackage{amsmath,amssymb,amsfonts}
\usepackage{graphicx,textcomp,bm,etoolbox,array}
\usepackage[hidelinks,bookmarks=false]{hyperref}
\hypersetup{
  pdftitle={Invertible Graph Aggregation Can Increase ReLU Readout Width},
  pdfauthor={Hiroyasu Usami, Masanori Tsukamoto, Kohei Hashiguchi},
  pdfsubject={IEEE Access manuscript: finite-width shared nodewise classification after graph aggregation},
  pdfkeywords={Graph neural networks, graph aggregation, readout capacity, ReLU networks, node classification}
}
\makeatletter
\AtBeginDocument{\DeclareMathVersion{bold}
\SetSymbolFont{operators}{bold}{T1}{times}{b}{n}
\SetSymbolFont{NewLetters}{bold}{T1}{times}{b}{it}
\SetMathAlphabet{\mathrm}{bold}{T1}{times}{b}{n}
\SetMathAlphabet{\mathit}{bold}{T1}{times}{b}{it}
\SetMathAlphabet{\mathbf}{bold}{T1}{times}{b}{n}
\SetMathAlphabet{\mathtt}{bold}{OT1}{pcr}{b}{n}
\SetSymbolFont{symbols}{bold}{OMS}{cmsy}{b}{n}
\renewcommand\boldmath{\@nomath\boldmath\mathversion{bold}}}
\patchcmd{\@maketitle}{\vss\doifont Digital Object Identifier\space\@doi\vss}{}{}{\PackageError{access-migration}{DOI patch failed}{}}
\let\access@titlepage\ps@titlepage
\def\ps@titlepage{\access@titlepage
  \def\@oddfoot{\hfil{\footerpagefont\thepage}}
  \def\@evenfoot{{\footerpagefont\thepage}\hfil}}
\let\access@headings\ps@headings
\def\ps@headings{\access@headings
  \def\@oddfoot{\hfil{\footerpagefont\thepage}}
  \def\@evenfoot{{\footerpagefont\thepage}\hfil}}
\ps@headings
\patchcmd{\IEEEbiographynophoto}{\vskip 4\baselineskip plus 1fil minus 0\baselineskip}{\vskip 2\baselineskip}{}{\PackageError{access-migration}{Biography spacing patch failed}{}}
\def\@IEEEBIOskipN{1\baselineskip}
\expandafter\patchcmd\csname\string\IEEEbiography\endcsname
{\vskip \@IEEEBIOskipN plus 1fil minus 0\baselineskip}
{\vskip \@IEEEBIOskipN}{}{\PackageError{access-layout}{Biography glue patch failed}{}}
\makeatother
\newcommand{\relu}{\operatorname{ReLU}}
\newcommand{\err}{\operatorname{err}}
\newcommand{\sr}{\operatorname{sr}}

\begin{document}
\history{}
\doi{}
\title{Invertible Graph Aggregation\\Can Increase ReLU Readout Width}
\author{\uppercase{Hiroyasu Usami}\authorrefmark{1},
\uppercase{Masanori Tsukamoto}\authorrefmark{2}, and
\uppercase{Kohei Hashiguchi}\authorrefmark{2}}
\address[1]{Department of Computer Science, Faculty of Engineering, Chubu University, Kasugai, Japan (e-mail: \href{mailto:usami@fsc.chubu.ac.jp}{usami@fsc.chubu.ac.jp})}
\address[2]{Department of Systemic Management for Dentistry, Kagoshima University Hospital, Kagoshima, Japan (e-mails: \href{mailto:tsukamoto@dent.kagoshima-u.ac.jp}{tsukamoto@dent.kagoshima-u.ac.jp} (M. Tsukamoto); \href{mailto:hashiguchi@dent.kagoshima-u.ac.jp}{hashiguchi@dent.kagoshima-u.ac.jp} (K. Hashiguchi))}
\tfootnote{This work has been submitted to the IEEE for possible publication. Copyright may be transferred without notice, after which this version may no longer be accessible.}
\markboth{Usami \headeretal: ReLU Readout Width After Graph Aggregation}{Usami \headeretal: ReLU Readout Width After Graph Aggregation}
\begin{abstract}
An invertible graph aggregation preserves the complete feature table, but can increase the size of a classifier that reads one node at a time. For every even $N\ge4$, we construct a labeled tree with $2N$ vertices and maximum degree three. A shared one-hidden-layer rectified linear unit (ReLU) classifier needs one hidden unit on the raw scalar features and exactly $N$ units after one lazy neighborhood mean. All aggregate values are distinct, and the mixing matrix has condition number at most 8. We determine the complete error curve for heads with a free affine term. The width separation persists under scalar perturbations. Under a specified bounded uniform-noise model with one nuisance coordinate, every one-unit aggregate head has expected error at least $1/12$, while a fixed $N$-unit head has zero error. Controlled fits compare trained heads with these explicit feasible scores. On Roman-empire and Tolokers, aggregation improves fitted prediction, and retaining raw features helps at the largest tested parameter budget. Training-only diagnostics include conditional output-layer optimality bounds for eight fixed runs.
\end{abstract}
\titlepgskip=-21pt
\maketitle

\section{Introduction}
\PARstart{G}{raph} aggregation changes the feature vectors presented to a nodewise classifier. An invertible aggregation allows the original feature table to be recovered from the complete output table. A classifier that receives only one node's output vector may nevertheless need a more complex decision function to recover that node's label.

Our readout is a shared one-hidden-layer ReLU classifier: the same score function is applied to each vertex's current feature vector. Its width is the number of hidden units. For the same fixed graph, features, and labels, we choose the head separately before and after aggregation and determine the smallest width permitting zero error. We also determine the error attainable at each width.

On a degree-three tree with $2N$ vertices, averaging each feature equally with its self-inclusive neighborhood mean raises the minimum width from 1 to exactly $N$. Every aggregate row remains distinct, each self-weight is at least $5/8$, and the mixing matrix has condition number at most 8. Allowing a free affine term gives the exact error at every width. Both conclusions persist throughout an explicit scalar neighborhood.

\begin{figure*}[!t]
\centering
\includegraphics[width=\textwidth]{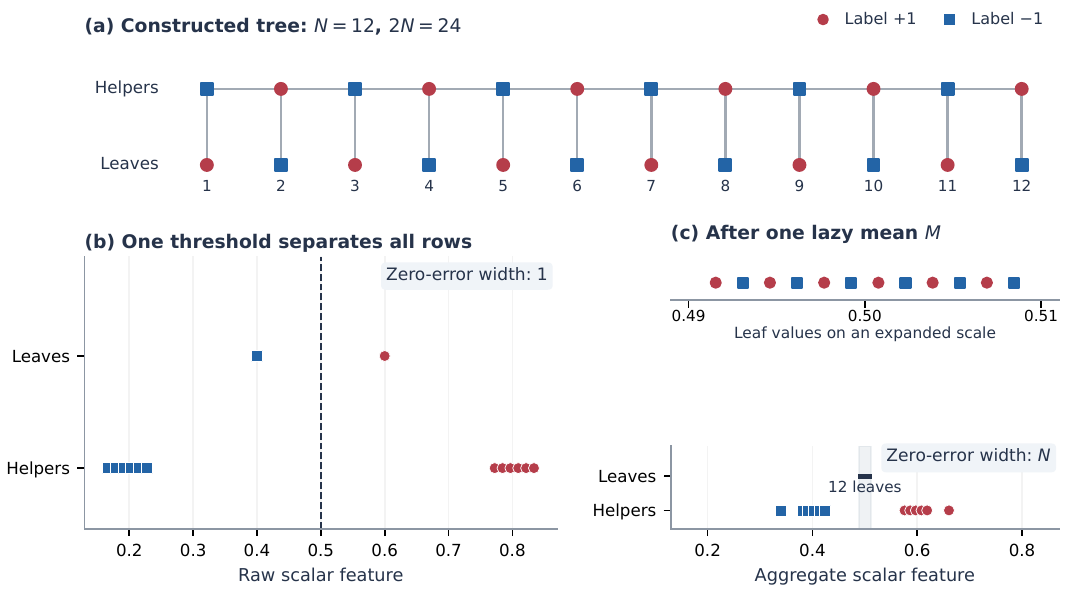}
\caption{Readout width on the degree-three tree, shown for $N=12$. Helpers form a path, each with one leaf. Circles and squares denote labels $+1$ and $-1$. Panel (c) separates the full aggregate axis from the expanded leaf axis; the shaded band locates the expanded interval. Raw features need one ordinary ReLU unit. One lazy mean creates $N+1$ label transitions and requires $N$ units, while retaining distinct aggregate rows and condition number at most 8.}\label{fig:mechanism}
\end{figure*}

The proof realizes classical sign-transition bounds through a local average, with conditioning controlled independently of $N$ (Fig.~\ref{fig:mechanism}). A full-rank extension examines global feature conditioning. A specified bounded uniform-noise model retains population width $N$ and, with one nuisance coordinate, gives one-unit error at least $1/12$. Controlled fits compare trained heads with feasible scores; two public benchmarks compare raw, mixed, and concatenated width curves under common optimization and selection rules.

\subsection{Relation to existing theory}
Baranwal, Fountoulakis, and Jagannath~\cite{baranwal2021} analyze linear separability and logistic-loss generalization in a Gaussian contextual stochastic block model. Their Theorems 1 and 2 impose explicit graph-density and class-contrast assumptions. Ghogho~\cite{ghogho2024} analyzes weighted-sum and self/neighbor-concatenation aggregation using statistical class models and degree-dependent linear classifiers. Wang \emph{et al.}~\cite{wang2024} compare separability before and after graph convolution in a heterophilous stochastic block model. These analyses use statistical graph and feature models; here the labeled table is fixed. Telgarsky~\cite{telgarsky2016} develops classical oscillation tools for neural-network lower bounds; our transition count is a one-dimensional instance of this approach.

H2GCN separates ego and neighbor representations and retains intermediate features~\cite{zhu2020heterophily}; informative class-dependent neighborhoods can support GCN classification at low homophily~\cite{ma2022homophily}. Our opposite-label tree quantifies exact readout-width growth after an invertible mean of a separable table.

Rank-based oversmoothing diagnostics already require care: Zhang \emph{et al.}~\cite{zhang2026} discuss limitations of rank-based diagnostics and distinguish representation rank from predictive performance. Their diagnostic question differs from the minimum-width comparison here. Keriven~\cite{keriven2026} studies backward oversmoothing and spurious stationary points in deep graph networks.

Nikolentzos, Kelesis, and Nakis~\cite{nikolentzos2026} construct invertible GNN layers. Their Proposition 3.1 already gives an invertible aggregation with coincident outputs at vertices distinguished by the one-dimensional Weisfeiler--Leman (1-WL) test. Our aggregate outputs all remain distinct; the cost is the exact width needed to read labels that were one-unit separable before aggregation. Pham and Garg~\cite{pham2024}, in Proposition 6 and Table 1, compare the depth and parameter requirements of message-passing and feedforward architectures for representing functions of input vectors using auxiliary graphs.

Chen, Wang, and Xiong~\cite{chen2023} study GCN activation regions as the whole input table varies. Chung and Sun~\cite{chung2026} give graph-Barron approximation results for scalar functions of a whole graph signal, using learned graph convolutions and vertex-indexed parameters. These whole-table function classes differ from a shared row readout.

\section{Graph, Features, and Readout Classes}\label{sec:setup}
Let $N\ge4$ be even. The helpers $h_1,\ldots,h_N$ form a path, and leaf $c_i$ is adjacent only to $h_i$. This is a finite undirected simple tree with $2N$ vertices and maximum degree three. Write $\mathcal{N}[v]$ for the closed neighborhood, including $v$, and set $D_v=\deg(v)+1$. Define
\begin{equation}\label{eq:mix}
 P_{vw}=\frac{\mathbf{1}\{w\in\mathcal{N}[v]\}}{D_v},
 \qquad M=\frac{I+P}{2}.
\end{equation}
Here $I$ is the $2N\times2N$ identity, $\mathbf{1}\{\cdot\}$ denotes an indicator, and $\mathbf{1}_k$ is the all-ones column of length $k$. A superscript $\mathsf T$ denotes transpose. Both $P$ and $M$ are row-stochastic: their entries are nonnegative and each row sums to one. For scalar features $F\in\mathbb{R}^{2N}$, the aggregate is $G=MF$.

A shared classifier receives only $G_v$ at vertex $v$. Labels are $y_v\in\{-1,+1\}$. The function $\operatorname{pred}(a)$ returns the sign of $a$, with one fixed label assigned to zero throughout. For either scalar or vector inputs, define
\begin{equation}\label{eq:error}
 \err(s;G,y)=\frac{1}{2N}\sum_v
 \mathbf{1}\{\operatorname{pred}(s(g_v))\ne y_v\}.
\end{equation}
For a scalar table, $g_v=G_v$; for a multivariate table, row $v$ is $g_v^{\mathsf T}$. All $2N$ graph vertices are assessed with equal weight. The classifier receives no vertex index, other rows, graph context, or raw-feature bypass.

Let $m$ be a nonnegative integer. An ordinary width-$m$ scalar score has one hidden ReLU layer and an output bias, and belongs to
\begin{equation}\label{eq:ordinary}
 \mathcal{O}_m=\left\{b+\sum_{r=1}^m a_r\relu(u_rz+v_r)\right\}.
\end{equation}
Here $\relu(q)=\max\{q,0\}$. All coefficients are real; input weights may be negative. Width zero permits only a constant. We also use the larger class
\begin{equation}\label{eq:hinge}
 \mathcal{H}_m=\left\{b+az+\sum_{r=1}^m c_r\relu(z-k_r)\right\}.
\end{equation}
The term $az$ is a direct linear connection from input to output and does not count toward $m$; thus $\mathcal{H}_0$ contains every affine score. Converting left-facing hinges shows $\mathcal{O}_m\subseteq\mathcal{H}_m$. The exact error curve below is for $\mathcal{H}_m$; the exact zero-error width is for $\mathcal{O}_m$.

Vector norms are Euclidean unless specified. The infinity norm is the largest absolute entry. Matrix $2$-norms are spectral norms, and $\|B\|_F$ is the square root of the sum of squared entries. Write $\sigma_{\max}(B)$ and $\sigma_{\min}(B)$ for its largest and smallest singular values. For a full-column-rank matrix, $\kappa_2(B)=\sigma_{\max}(B)/\sigma_{\min}(B)$; its stable rank is $\sr(B)=\|B\|_F^2/\|B\|_2^2$.

\section{Exact Readout Width}\label{sec:scalar}
The raw features admit a threshold separator. Their aggregate leaf values will instead alternate in label order. For $i=1,\ldots,N$, set
\begin{equation}\label{eq:features}
\begin{aligned}
 \sigma_i&=(-1)^{i+1},&
 t_i&=\frac{2i-N-1}{100(N+1)},\\
 F_{c_i}&=\frac12+\frac{\sigma_i}{10},&
 F_{h_i}&=\frac12+4t_i-\frac{3\sigma_i}{10}.
\end{aligned}
\end{equation}
Assign $y_{c_i}=\sigma_i$ and $y_{h_i}=-\sigma_i$, equivalently $y_v=\operatorname{sign}(F_v-1/2)$. Every graph edge joins opposite labels. The center $1/2$ is the raw decision threshold, and the leaf amplitude $1/10$ sets the minimum signed margin $y_v(F_v-1/2)$. At a leaf, $M$ gives weights $3/4$ and $1/4$ to the leaf and helper. The helper coefficients $4t_i$ and $-3\sigma_i/10$ therefore cancel the label offset and leave the aggregate $1/2+t_i$. The denominator 100 ensures $|t_i|<0.01$, keeping all features strictly inside the unit interval.

\noindent\textbf{Theorem 1 (exact scalar separation).}
For the graph, operator, features, and labels above:
\begin{enumerate}
\item $F\in(0.16,0.84)^{2N}$. The unit-slope raw score $z-1/2$ has signed margin at least $1/10$ at every vertex. The minimum ordinary raw width is 1.
\item $M$ is invertible and $\kappa_2(M)\le8$, independently of $N$.
\item All $2N$ entries of $G=MF$ are distinct.
\item The minimum ordinary aggregate width for zero error is exactly $N$.
\item For every integer $m\ge0$,
\begin{equation}\label{eq:risk}
 \min_{s\in\mathcal{H}_m}\err(s;G,y)
 =\frac{\left\lceil (N-m)_+/2\right\rceil}{2N}.
\end{equation}
Here $x_+=\max\{x,0\}$. The right side is also a lower bound for $s\in\mathcal{O}_m$.
\end{enumerate}

Appendix~\ref{app:scalar} bounds the conditioning, counts label transitions, and constructs an attaining ordinary $N$-unit score.

\begin{figure*}[!t]
\centering
\includegraphics[width=0.95\textwidth]{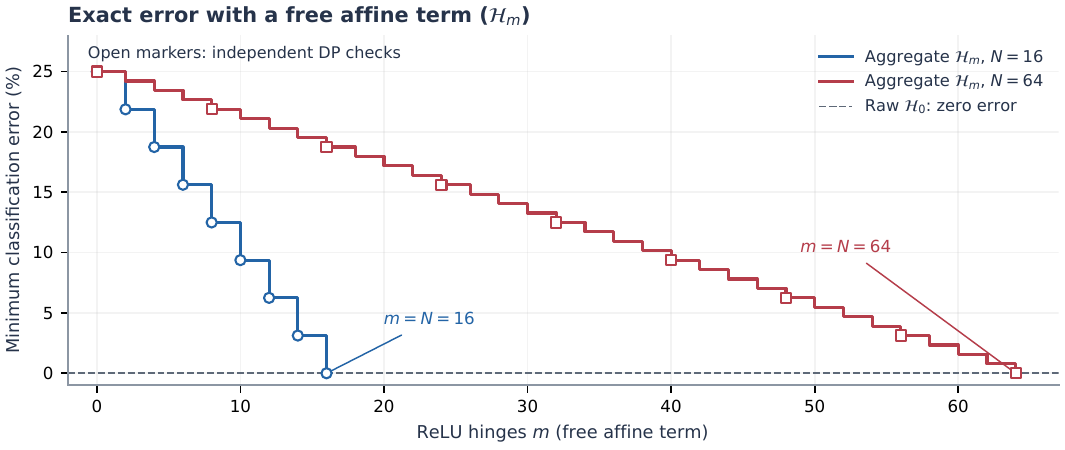}
\caption{Exact minimum error in $\mathcal{H}_m$, which includes an uncounted affine term. Open markers are independent dynamic-programming checks. The raw affine separator has zero error already in $\mathcal{H}_0$. For ordinary heads $\mathcal{O}_m$, the aggregate curve is a lower bound and the exact zero-error width is $N$; the raw ordinary width is 1.}\label{fig:risk}
\end{figure*}

The aggregate leaf values are $\xi_i=1/2+t_i$, equally spaced by $h=1/[50(N+1)]$. The helper values lie in two outer blocks. In increasing output order, the labels are a negative helper block, the alternating leaf labels $(+,-,\ldots,+,-)$, and a positive helper block. There are exactly $N+1$ transitions. A score in $\mathcal{H}_m$ makes at most $m+1$ transitions, and each mistaken label can remove at most two required transitions. This yields the lower bound in~\eqref{eq:risk}; explicit interpolants attain it. Figure~\ref{fig:risk} displays this class-optimal curve.

\subsection{Stability under scalar input perturbations}
\noindent\textbf{Proposition 1 (positive-radius scalar neighborhood).}
For the fixed labels of Theorem 1, let $\widetilde F=F+E$ with
\begin{equation}\label{eq:radius}
 \|E\|_\infty\le\eta_N=\frac{h}{4}
 =\frac{1}{200(N+1)}.
\end{equation}
Then $\widetilde F\in[0,1]^{2N}$, and the raw unit-slope margin is at least $1/10-\eta_N$. The order of all aggregate entries is unchanged. The ordinary aggregate zero-error width remains exactly $N$, and~\eqref{eq:risk} holds with $G$ replaced by $M\widetilde F$.

Appendix~\ref{app:stability} proves the claim using the exact minimum aggregate gap $h$ and Lipschitz interpolants. The neighborhood radius is positive for each $N$ and decreases with $N$. The ordinary width-$N$ construction can be adjusted to the perturbed leaf positions.

\subsection{A population extension under common nuisance overlap}
\label{sec:population_overlap}
The width increase also persists under fresh feature noise: the primary coordinate stays near its constructed value while additional coordinates vary. Let $V$ be the $2N$ vertices of Theorem~1 and put $I_v=(G_v-\eta_N,G_v+\eta_N)$, using the clean aggregate $G$ and radius in~\eqref{eq:radius}.
Let $q\ge1$ count these additional nuisance coordinates, so the input dimension is $d=q+1$. At vertex $v$, let $A_v\in I_v$ almost surely be the random primary coordinate and $Z_v\in\mathbb R^q$ the nuisance vector. Suppose $Z_v$ has Lebesgue density $p_v$ and
\begin{equation}\label{eq:population_overlap}
 \omega=\int_{\mathbb R^q}\min_{v\in V}p_v(z)\,dz>0.
\end{equation}
The overlap $\omega$ is the probability mass common to all nuisance densities. Dependence within a row or between vertices is allowed. Write $\mathcal O_m^{q+1}$ for the ordinary multivariate width-$m$ class, with $q+1$ inputs, and define risk averaged equally over vertices by
\begin{equation}\label{eq:population_risk}
 \mathcal R(s)=\frac1{2N}\sum_{v\in V}
 \mathbb E\,\mathbf{1}\{\operatorname{pred}(s(A_v,Z_v))\ne y_v\}.
\end{equation}

\noindent\textbf{Corollary 1 (population readout width).}
For every $s\in\mathcal O_m^{q+1}$,
\begin{equation}\label{eq:population_width_bound}
 \mathcal R(s)\ge
 \frac{\omega}{2N}\left\lceil\frac{(N-m)_+}{2}\right\rceil.
\end{equation}
A fixed $N$-unit head ignoring $Z_v$ has zero risk. Thus the minimum population zero-error width is exactly $N$.

Appendix~\ref{app:population_overlap} proves the result using the slice-integration principle of Telgarsky~\cite[Theorem~3.12]{telgarsky2016}. For a fitted head, evaluation noise is independent of the fit.

\paragraph{Uniform noise with a quantitative error bound.}
Independently perturb every raw primary coordinate uniformly on $(-\eta_N,\eta_N)$ and add $q$ independent raw nuisance coordinates uniform on $(1/2-\beta,1/2+\beta)$, where $0<\beta<1/2$. After aggregation, the row distributions have full-dimensional densities and
\begin{equation}\label{eq:uniform_noise_risk}
 \omega\ge3^{-q},\qquad
 \mathcal R(s)\ge\frac{3^{-q}}{2N}
 \left\lceil\frac{(N-m)_+}{2}\right\rceil.
\end{equation}
With one nuisance coordinate, every width-1 head therefore has expected error at least $1/12$ for every even $N\ge4$. The fixed $N$-unit head has zero risk, as does the raw one-unit separator. The overlap bound is independent of $N$ and $\beta$; the primary radius $\eta_N$ still decreases with $N$.

\section{Global Feature Conditioning and a Hard Subset}\label{sec:highdim}
The scalar construction extends to a full-column-rank table. Let $N=2^r$ with an integer $r\ge3$, set $\gamma=1/10$, and write the binary digits of $k=i-1$ as $k_b$, $b=0,\ldots,r-1$. For a subset $J$ of bit positions, let
\[
 W_J(k)=(-1)^{\sum_{b\in J}k_b}.
\]
Write $\sigma=(\sigma_i)_{i=1}^N$, $t=(t_i)_{i=1}^N$, and $W_J=(W_J(i-1))_{i=1}^N$. Distinct Walsh characters are orthogonal. In particular, $\sigma=W_{\{0\}}$ and
\[
 t_i=-\frac{\sum_{b=0}^{r-1}2^bW_{\{b\}}(i-1)}{100(N+1)}.
\]
Choose $d-1$ distinct characters with $|J|\ge2$, where $1\le d\le N-r$. In leaf/helper row order, define the centered columns
\begin{equation}\label{eq:walsh}
\begin{aligned}
 X_0&=\begin{pmatrix}\gamma\sigma\\4t-3\gamma\sigma\end{pmatrix},&
 X_J&=\begin{pmatrix}a_J\\-3a_J\end{pmatrix},\quad a_J=\gamma W_J,\\
 F&=\tfrac12\mathbf{1}_{2N}\mathbf{1}_d^{\mathsf T}+X.
\end{aligned}
\end{equation}
Column 0 is the primary coordinate; each chosen $J$ supplies one nuisance coordinate.

\noindent\textbf{Proposition 2 (full-rank extension).}
The columns of $X$ have zero global mean. The primary coordinate retains every label and the raw margin of Theorem 1. All coordinates of $F$ lie in $(0.16,0.84)$. Moreover,
\begin{equation}\label{eq:xcondition}
 \operatorname{rank}(X)=d,\quad
 \kappa_2(X)\le\sqrt{1.1},\quad \sr(X)\ge d/1.1.
\end{equation}
With $H=I-\mathbf{1}_{2N}\mathbf{1}_{2N}^{\mathsf T}/(2N)$ and $Y=HMX$,
\begin{equation}\label{eq:ycondition}
 \operatorname{rank}(Y)=d,\quad
 \kappa_2(Y)\le8\sqrt{1.1},\quad \sr(Y)\ge d/70.4.
\end{equation}
All aggregate rows are distinct. For a shared multivariate ordinary width-$m$ head, at least
\begin{equation}\label{eq:multirisk}
 \left\lceil (N-m-2)_+/2\right\rceil
\end{equation}
leaves are misclassified. Its error on all vertices is at least this count divided by $2N$, and its minimum zero-error width lies in $[N-2,N]$.

Here a multivariate ordinary head replaces $u_rz$ in~\eqref{eq:ordinary} by $u_r^{\mathsf T}z$, $u_r\in\mathbb{R}^d$. The lower bound follows from the leaf geometry:
\begin{equation}\label{eq:leafline}
 g_{c_i}=(1/2+t_i,1/2,\ldots,1/2)^{\mathsf T}.
\end{equation}
The nuisance coordinates cancel at leaves, so every such head restricts to a univariate score there. Thus the example separates whole-table conditioning from the collinearity of its hard subset. The denominator $70.4=8^2\cdot1.1$ comes from the squared condition bound. Appendix~\ref{app:highdim} supplies the calculations, including centering after a mean that is not necessarily doubly stochastic.

\subsection{An additional feature direction can remove the obstruction}
\noindent\textbf{Proposition 3 (ambient perturbation counterexample).}
If $d\ge2$, choose one nuisance coordinate and add $\epsilon\sigma_i$ to both its raw leaf and paired helper, where $0<\epsilon\le0.01$. Leave all other coordinates unchanged. All features remain in $[0,1]$, and an ordinary three-unit head classifies every aggregate row correctly.

Writing the primary and chosen nuisance aggregate coordinates as $z_0,z_1$, the score is
\begin{equation}\label{eq:three}
 \relu(z_1)-\frac12
 +10\relu(z_0-0.52)-10\relu(0.48-z_0).
\end{equation}
The cutoffs $0.48$ and $0.52$ lie outside the leaf interval $(0.49,0.51)$ and between the leaves and helper blocks. The gain 10 makes the helper correction dominate the nuisance coordinate, as shown in Appendix~\ref{app:highdim}. Full rank persists for sufficiently small $\epsilon$, and conditioning converges to that of the unperturbed table as $\epsilon\downarrow0$. Consequently the high-dimensional lower bound is not stable to arbitrary ambient perturbations. The scalar neighborhood in Proposition 1 has no additional coordinate in which to carry this label signal.

\section{Exact Construction Checks}\label{sec:checks}
\subsection{Construction checks}
The construction checker \textnormal{verify\_capacity\_construction.py} checks $N=4,8,16,32,64$, corresponding to 8 through 128 vertices. Graph outputs, labels, explicit scalar scores, and perturbation examples use rational arithmetic. A dynamic program minimizes errors over binary prediction sequences with at most $m+1$ changes, without using~\eqref{eq:risk}. All 129 checked profile points agree with the formula. The analytic interpolants establish attainability by $\mathcal{H}_m$.

Walsh orthogonality is checked with integer products, and the ambient perturbation example uses rational arithmetic. High-dimensional matrix products, centering, and singular-value diagnostics use float64. An independent Fraction-only script checks the adjusted ordinary $N$-unit score on 328 scalar perturbations, including all 256 coordinatewise boundary perturbations at $N=4$. A further rational-arithmetic check confirms nonnegative endpoint scores and strict correctness inside the population intervals at eight tree sizes. The checker scripts and figure-generation code accompany the manuscript.

\section{Controlled Training Study}\label{sec:controlled}
For the clean scalar construction, the theorem specifies the minimum zero-error width and an explicit classifier attaining it. We use these as references for trained heads, then vary the mixing strength, scalar perturbations, and additional feature directions.

\subsection{Protocol}
We use $N=8,16$ and ordinary binary heads of widths $m\in\{1,3,8,16,32\}$. The grid includes the one- and three-unit controls, both exact theorem widths, and width 32. The three inputs are the raw table $F$, its aggregate $M_\alpha F$, and their concatenation $[F,M_\alpha F]$, where $M_\alpha=(1-\alpha)I+\alpha P$. Concatenation retains a raw-feature bypass and increases the parameter count at a given width. Every input coordinate receives the same fixed transform $z\mapsto100(z-1/2)$. This rescales the narrow leaf interval without using labels or fitted coordinate statistics.

At $\alpha=1/2$, we evaluate six conditions: the exact scalar table; independent uniform raw scalar perturbations of radii $\eta_N$ and $4\eta_N$; two or eight independent raw nuisance coordinates, each uniform on $[0.49,0.51]$; and the Walsh-based extra coordinate in Proposition 3 with label signal $\epsilon=0.001$ and independent jitter of radius $\epsilon/4=0.00025$. The last condition is a supervised control for the perturbation counterexample. Four further conditions use the unchanged exact table at $\alpha=0.25,0.49,0.51,0.75$. The features are not redesigned for these mixing strengths.

Each of three paired seeds controls feature sampling and initialization. We generate one training, eight validation, and sixteen test realizations at the same $2N$ labeled vertices. Noisy realizations assess feature transfer on this fixed graph; noiseless realizations repeat one table for optimization calibration. Seed variation combines sampling and initialization.

Each candidate uses 400 full-batch Adam updates followed by at most 60 L-BFGS iterations, in float64, with binary cross-entropy and no weight decay. Adam learning rates are $0.01$ and $0.03$; L-BFGS uses rate 1 with a strong-Wolfe line search. Raw and aggregated heads have identical initial states at each matched width, seed, and rate. We select checkpoints and rates by validation cross-entropy before computing that cell's test scores. The complete grid comprises 1,800 candidate fits and 900 selected fits. We record the training-gradient $\ell_2$ norm at the selected checkpoint and the final trajectory endpoint, using $10^{-4}$ as a diagnostic threshold. The objective-plateau diagnostic uses only the final L-BFGS checkpoints. Cross-entropy on a separable table need not have a finite minimizer; these diagnostics do not certify a global optimum.

\subsection{Results and interpretation}
Raw and concatenated inputs reach zero training and test error in all 300 selected fits for each arm, including width 1. On the exact aggregated table, the width-1 error is 25\% for both tree sizes. At the sufficient width $m=N$, the learned error still averages 22.92\% for both $N=8$ and $N=16$, whereas the fixed explicit $N$-unit score has zero error. Of the fifteen exact-table aggregated fits with $m\ge N$, fourteen retain errors. Increasing to width 32 gives mean errors of 8.33\% and 22.92\% at $N=8$ and $N=16$, respectively. Under the prescribed training and selection budget, the fitted heads do not reach the available zero-error solution.

\begin{figure*}[!t]
\centering
\includegraphics[width=\textwidth]{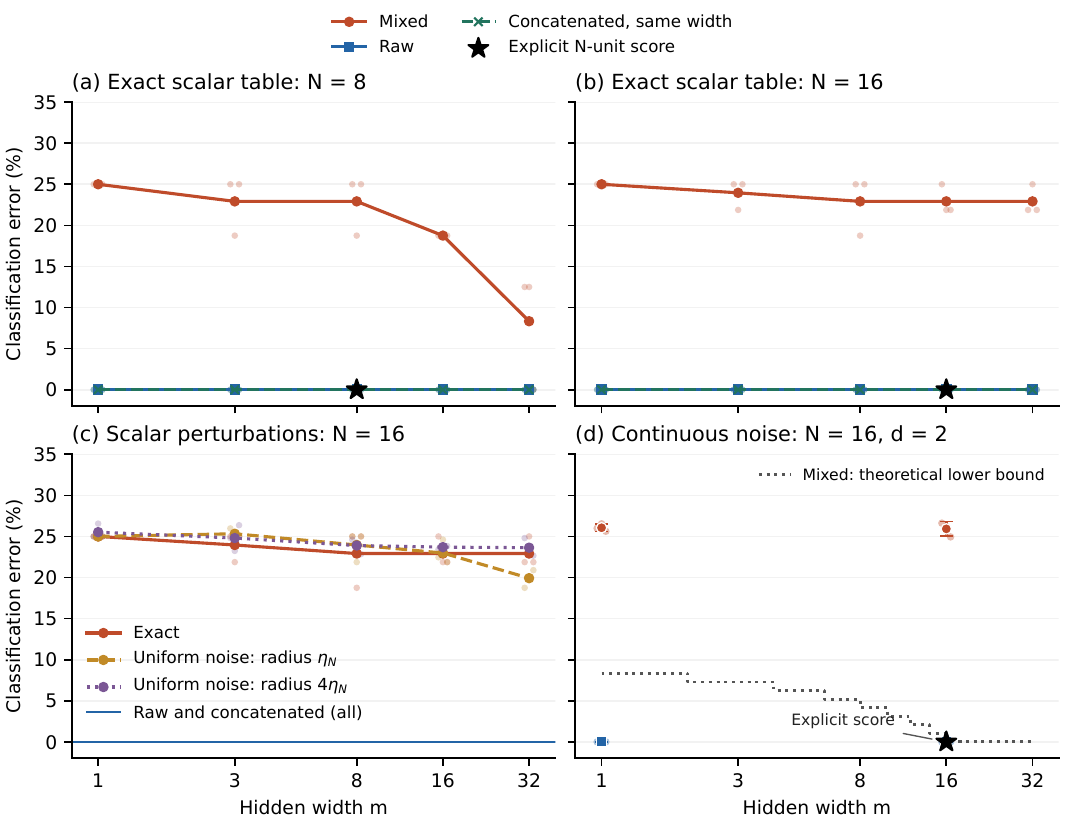}
\caption{Controlled readout fits. Large symbols show means over three paired data/initialization seeds; faint points show individual seeds. Panels (a)--(c) use sixteen evaluation realizations per seed; (a) and (b) repeat the noiseless training table. Panel (d) uses 4,096 fresh whole-graph realizations per seed with continuous primary and nuisance noise. Only widths 1 and 16 are fitted there; bars show seed SD. Its dotted curve is the population lower bound from~\eqref{eq:uniform_noise_risk}. Black stars show fixed explicit $N$-unit scores.}\label{fig:controlled}
\end{figure*}

Scalar perturbations within the proved radius retain the aggregate difficulty for the trained heads: at $N=16$, the mean error decreases from 25.00\% at width 1 to 19.92\% at width 32. Beyond that radius, adjacent leaf-value distributions overlap. The resulting errors combine noise ambiguity with fitting error and cannot isolate a capacity barrier. Independent nuisance directions also leave substantial evaluation error in the aggregated arm. By contrast, the explicitly label-bearing coordinate gives zero training and evaluation error from width 3 onward for both tree sizes. Corollary~1 gives a corresponding population statement: common nuisance overlap leaves a positive error lower bound for narrow fixed heads, whereas the label-bearing direction separates opposite-label leaf supports. These extra-coordinate controls hold the primary coordinate fixed; their per-fit results remain in the companion tables.

The unchanged table is easy at $\alpha=0.25$: every selected arm reaches zero error even at width 1. At $\alpha=0.49$ and $0.51$, the scalar label order retains many transitions while the measured condition numbers remain below 3.5. The difficult order therefore persists at the tested values on both sides of $1/2$. The $\alpha=0.75$ control is poorly conditioned and falls outside the uniformly bounded-conditioning example (Table~\ref{tab:alpha}).

\begin{table}[!t]
\caption{Mixing-strength controls on unchanged noiseless features. Transition counts and aggregated-head mean width-1 errors are listed as $N=8/N=16$. Condition numbers are numerical diagnostics of the actual $M_\alpha$.}
\label{tab:alpha}
\centering\small\setlength{\tabcolsep}{3pt}
\begin{tabular}{rrrrr}
\hline
$\alpha$ & $\kappa_2$, $N=8$ & $\kappa_2$, $N=16$ & Transitions & Error (\%)\\
\hline
0.25 & 1.54 & 1.54 & 1/1 & 0.00/0.00 \\
0.49 & 3.12 & 3.18 & 7/13 & 25.00/18.75 \\
0.50 & 3.25 & 3.33 & 9/17 & 25.00/25.00 \\
0.51 & 3.40 & 3.49 & 7/13 & 25.00/25.00 \\
0.75 & 49.16 & 1329.67 & 7/15 & 33.33/33.33 \\
\hline
\end{tabular}
\end{table}

The selected-state gradient threshold is met in 711 of 900 fits; the final-state threshold is met in 766, and the final objective-plateau diagnostic in 863. These counts concern different states and criteria. An independent checker reloads saved checkpoints, recomputes forward scores, verifies paired initializations and validation selection, and reconciles every reported aggregate with the per-fit records.

\subsection{Fresh continuous feature noise}
We add one prespecified condition with $N=16$, one nuisance coordinate ($d=2$), and $\beta=0.01$, using the independent uniform primary and nuisance perturbations in~\eqref{eq:uniform_noise_risk}. Raw and aggregated heads have widths 1 or 16. Three paired seeds each provide one training realization and eight validation realizations. The input transform and optimizer budget match the preceding study, with Adam rate 0.01. Validation cross-entropy selects one checkpoint per fit. After all twelve choices are fixed, we generate 4,096 independent whole-graph test realizations per seed.

Figure~\ref{fig:controlled}(d) shows test errors of $26.05\pm0.46\%$ and $25.92\pm0.87\%$ for aggregated widths 1 and 16, respectively (mean $\pm$ sample SD over seeds). Their conditional Monte Carlo standard errors are 0.055--0.095 percentage points; vertices are averaged within each realization. Raw heads and the fixed 16-unit aggregate score make no errors. All selected aggregate width-16 heads are Adam checkpoints; their mean training error is 19.79\%. The gap from the feasible population score concerns this fitting-and-selection procedure, including transfer to fresh noise. The dotted curve gives the analytical lower bound in~\eqref{eq:uniform_noise_risk}.

\section{Public-Benchmark Readout Study}\label{sec:public}
The controlled study fixes labels on a constructed graph. We next ask how the width dependence of a trained readout changes after one lazy mean on two public benchmarks.

\subsection{Data and representations}
We use Roman-empire and Tolokers from Platonov \emph{et al.}~\cite{platonov2023}, with all ten official 50/25/25 train/validation/test splits. Roman-empire has 22,662 nodes, 300 provided features, 18 classes, and 32,927 undirected edges. Tolokers has 11,758 nodes, 10 features, two classes, and 519,000 undirected edges. The dataset pair and split indices were fixed before the present test evaluation.

Let $Z$ contain the provided features standardized using each split's training-node means and standard deviations. We symmetrize and deduplicate graph support, add one self-loop per vertex, and form the row-stochastic mean $P$. The aggregate is $G=(Z+PZ)/2$. This half-mean is invertible: its diagonal entry in every row exceeds the sum of its off-diagonal entries. No post-aggregation standardization is applied. Prepared arrays are stored in float32 and converted to float64 for optimization. All node features and graph edges are available transductively; only training labels enter the objective.

We compare $Z$, $G$, $[Z,G]$ at the same hidden width, and a parameter-budget version of $[Z,G]$. For input dimension $d$, $C$ classes, and width $m$, an ordinary multiclass head has $m(d+C+1)+C$ parameters. The budget version uses width $\lfloor m(d+C+1)/(2d+C+1)\rfloor$, so its parameter count does not exceed that of the width-$m$ raw head. These arms compare the raw-feature bypass at a common width and under a common upper parameter budget.

\subsection{Optimization, selection, and evaluation}
Every arm uses a shared \textnormal{Linear--ReLU--Linear} head, with both biases and no learned graph encoder, dropout, or batch normalization. Nominal widths are $8,32,128$, and candidate Adam learning rates are $0.003,0.01$. The seed is $4100$ plus the zero-based split index, paired between raw and aggregated heads at each width. We minimize training cross-entropy plus $\lambda\|\theta\|_2^2/2$, where $\theta$ contains every head parameter and $\lambda=10^{-4}$. This common weight-space penalty defines the fitted pipelines; changing feature coordinates or input dimension does not preserve an identical function-space norm constraint.

Every candidate uses full-batch float64: 300--1,200 Adam updates, with training-objective-based learning-rate reduction and a plateau rule of relative objective change below $10^{-4}$ over 100 updates, followed by up to twelve blocks of 50 L-BFGS iterations. A test-free CPU preflight increased the common L-BFGS cap from 200 to 600. Final endpoints pass the autograd-gradient diagnostic at training-objective gradient infinity norm $\le10^{-4}$; larger-gradient objective and parameter plateaus are flagged as stagnation, and capped nonpassing endpoints remain unresolved. Earlier validation-selected checkpoints have separate diagnostics. At ReLU kinks, a nonzero autograd gradient need not exclude generalized stationarity; a small gradient does not certify an optimum. The archive records settings and histories.

Within each candidate run, we select the checkpoint with lowest validation cross-entropy; the learning rate is selected by the same criterion. This surrogate is common to both datasets. The test-evaluation stage requires all 480 candidate fits and the 240 validation selections to be completed and sealed. We also evaluate the final training endpoints to show the effect of validation checkpoint selection.

The primary metrics follow the benchmark: accuracy for Roman-empire and ROC AUC for Tolokers. For each split, the width interaction is the mixed-minus-raw metric gap at width 128 minus that at width 8. We report all split values and summarize them by the mean and sample standard deviation. The ten splits overlap; their sample SD describes variation over these splits.

\subsection{Results}
\begin{figure*}[!t]
\centering
\includegraphics[width=\textwidth]{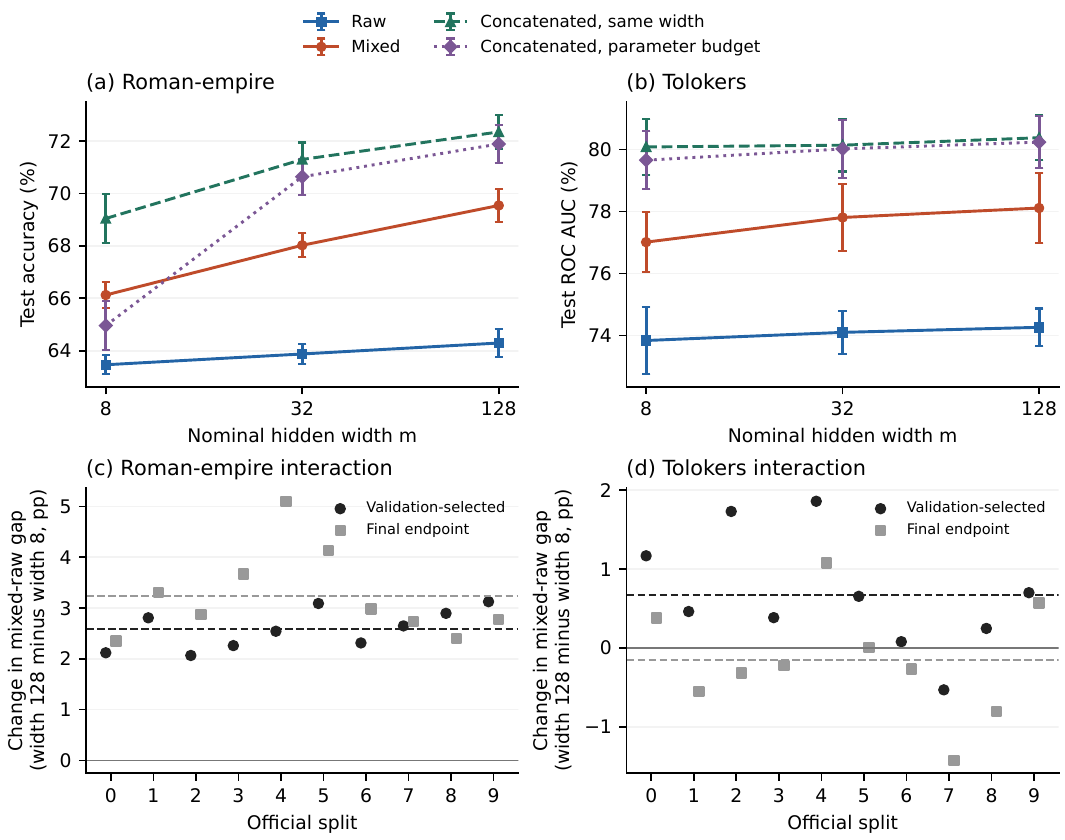}
\caption{Held-out public results over ten official splits. Top: validation-selected test metric, with the mean and sample SD across splits. Parameter-budget concatenation uses the reduced width defined in the text. Bottom: change in the mixed-minus-raw gap from width 8 to 128. Positive values mean that the mixed advantage grows with width. Circles and squares show validation-selected states and final endpoints of the same runs; dashed lines show their means.}\label{fig:public}
\end{figure*}

Aggregation improves validation-selected test performance at both endpoint widths on both datasets (Fig.~\ref{fig:public}). On Roman-empire, the mixed-minus-raw accuracy gap grows from $2.66$ percentage points at width 8 to $5.25$ at width 128. On Tolokers, the corresponding ROC AUC gaps are $3.17$ and $3.85$ points. Every split has a positive gap at these two widths.

\begin{table*}[!t]
\caption{Mixed-minus-raw gaps and width interaction, in percentage points (mean $\pm$ sample SD over ten overlapping splits). Roman-empire uses accuracy; Tolokers uses ROC AUC. ``Final'' denotes the final training checkpoint of the validation-chosen run.}
\label{tab:publicinteraction}
\centering\small\setlength{\tabcolsep}{8pt}
\begin{tabular}{llrrr}
\hline
Dataset & Checkpoint & Gap at width 8 & Gap at width 128 & Width interaction\\
\hline
Roman-empire & Selected & $+2.66\pm0.59$ & $+5.25\pm0.43$ & $+2.59\pm0.39$ \\
Roman-empire & Final & $+3.14\pm0.73$ & $+6.37\pm0.44$ & $+3.23\pm0.86$ \\
Tolokers & Selected & $+3.17\pm0.84$ & $+3.85\pm0.97$ & $+0.67\pm0.74$ \\
Tolokers & Final & $+3.27\pm0.71$ & $+3.11\pm0.70$ & $-0.16\pm0.72$ \\
\hline
\end{tabular}
\end{table*}

The selected-checkpoint width interaction is positive on all Roman-empire splits and nine Tolokers splits. The final-endpoint mean remains positive on Roman-empire, but changes sign on Tolokers (Table~\ref{tab:publicinteraction}). The Tolokers interaction therefore depends on checkpoint selection, even though its mixed-minus-raw endpoint gaps remain positive.

At nominal width 128, parameter-budget concatenation reaches $71.88\pm0.72$\% accuracy on Roman-empire and $80.24\pm0.84$\% ROC AUC on Tolokers, compared with $69.54\pm0.62$\% and $78.11\pm1.13$\% for mixed inputs. The budget heads use 65 and 72 actual units, respectively, with fewer parameters than their raw width-128 counterparts. At nominal width 8 on Roman-empire, budget concatenation instead falls below the mixed head ($64.96$\% versus $66.12$\%). The value of retaining raw features depends on the available head budget.

The common autograd-gradient threshold is met by 19 of 480 candidate final endpoints and 11 of the 240 chosen-run final endpoints. None of the 240 validation-selected states meets it; 230 of these states are Adam checkpoints. The eleven chosen-run passes are all at width 128 on Roman-empire (nine mixed and two same-width concatenated runs). The curves characterize the common fitting procedure; most final endpoints remain above the gradient threshold. Independent audits reconstruct every split and representation, recompute held-out predictions and ROC AUC, and check saved-state objectives and gradients against the records.

\subsection{Optimization diagnostics}
We examine the raw and mixed heads of widths 8 and 128 on official split 0 of each dataset, using the eight final checkpoints obtained with initial Adam learning rate $0.003$. From each checkpoint, we run three training-only continuations capped at 200 L-BFGS iterations: four restarted 50-iteration blocks, one optimizer with persistent history, and output-only fitting with hidden features fixed. The persistent optimizer produces a lower objective than restarted blocks in six cases; none of its endpoints meets the full-gradient threshold. Output-only fitting lowers the objective in all eight cases. Negative-gradient probes from the original checkpoints also find descent in seven.

For a fixed hidden layer, let $B_{\rm out}$ contain the output weights and biases. Let $J_{\rm out}$ be mean training cross-entropy plus the original ridge penalty, including the constant hidden penalty, and let $J_{\rm out}^\star$ be its minimum over $B_{\rm out}$. Strong convexity gives
\begin{equation}\label{eq:output_gap}
\begin{aligned}
 0&\le J_{\rm out}(B_{\rm out})-J_{\rm out}^\star\le\Delta_{\rm out},\\
 \Delta_{\rm out}&:=\frac{\|\nabla J_{\rm out}(B_{\rm out})\|_F^2}{2\lambda}.
\end{aligned}
\end{equation}
The softmax probabilities at the current logits define a feasible dual point with primal--dual gap exactly $\Delta_{\rm out}$. We evaluate this identity in all 32 saved states. Four of eight frozen-hidden endpoints meet the predeclared criterion $\Delta_{\rm out}\le10^{-5}$; no original endpoint does. Seven frozen-hidden endpoints meet the separate output-gradient infinity-norm threshold of $10^{-4}$. These certificates bound output-layer suboptimality for each fixed hidden layer. The companion table provides every gap and the float64 identity checks.

\section{Discussion}
One local mean can increase readout width while preserving the complete table and good conditioning. The construction attributes this increase to ordered label transitions; the controlled fits separately expose optimization gaps despite sufficient width. Capacity and optimization impose different downstream costs.

\subsection{Score normalization}
Norm-based capacity control~\cite{neyshabur2015} and margin-normalized generalization bounds~\cite{bartlett2017} provide context for reporting coefficient size and score normalization. The following estimate uses only the spacing of the constructed leaves.

At a fixed score margin, the required Lipschitz constant grows with $N$. Let $\rho>0$ and require $y_vs(G_v)\ge\rho$ at every vertex. Adjacent, oppositely labeled leaves have distance $h=1/[50(N+1)]$, so any $L$-Lipschitz shared score satisfies
\begin{equation}\label{eq:marginlip}
 Lh\ge2\rho,\qquad L\ge100\rho(N+1).
\end{equation}
For an ordinary head in~\eqref{eq:ordinary}, define $B=\sum_{r=1}^m|a_ru_r|$. Since $\operatorname{Lip}(s)\le B$,
\begin{equation}\label{eq:coeffbound}
 B\ge100\rho(N+1).
\end{equation}
On positive raw inputs, $\relu(z)-1/2$ has $B=1$ and margin at least $1/10$. We express $B$ in original input units. Positive score rescaling multiplies $\rho$ and $B$ equally, so their invariant ratio satisfies
\begin{equation}\label{eq:normalizedmargin}
 \frac{\rho}{B}\le\frac h2=2\eta_N.
\end{equation}
For the same fixed head and $\|E\|_\infty\le\eta_N$, row-stochasticity of $M$ gives
\begin{equation}\label{eq:fixedcertificate}
 y_vs((M(F+E))_v)\ge\rho-B\eta_N.
\end{equation}
A positive right-hand side certifies label preservation. Applying this certificate requires both the head's margin and its coefficient sum.

\subsection{Feature retention and fitted performance}
Independent nuisance coordinates can help fit one realization without improving fresh-noise prediction. The label-bearing direction admits a three-unit score, while a raw-feature bypass retains the original separator. These controls distinguish global feature conditioning from label-bearing directions.

Public results favor aggregation; the benefit of retention depends on head budget and checkpoint selection. Budget concatenation helps at width 128, but its narrow Roman-empire head loses to mixing, and the Tolokers interaction changes sign at the final endpoint. The construction supplies a separate worst-case boundary.

\section{Conclusion}
One lazy mean on a degree-three tree can preserve invertibility and good conditioning while increasing exact ordinary ReLU readout width from 1 to $N$. An explicit bounded uniform-noise model retains this population width separation and gives a quantitative error floor for narrow heads. Controlled fits distinguish available capacity from optimizer failure. On two public graphs, aggregation improves fitted prediction and feature retention helps at the largest tested budget; the width interaction depends on dataset and checkpoint selection. Readout width, retained-feature budgets, and optimization bounds describe different costs of the representation presented to a nodewise classifier.

\section*{Data and Code Availability}
Roman-empire, Tolokers, and their official splits are available from Platonov \emph{et al.}~\cite{platonov2023} at \url{https://github.com/yandex-research/heterophilous-graphs}. Code and the reproducibility archive are available in the \href{https://github.com/usamihiroyasu/invertible-graph-readout-width}{GitHub repository}; \href{https://github.com/usamihiroyasu/invertible-graph-readout-width/releases/tag/v1.0.0}{release v1.0.0} fixes the version accompanying this manuscript. The archive pins the upstream commit and hashes, verifies downloaded files, and contains construction checks, runners, fixed protocols, saved states, summaries, certificates, and plotting scripts. No clinical data are used.

\appendices
\section{Proof of Theorem 1}\label{app:scalar}
\subsection{Raw separator and condition number}
Leaves have values $0.4$ or $0.6$; helpers lie in $(0.16,0.24)$ or $(0.76,0.84)$. This proves the feature range and raw margin. Since all inputs are positive, $\relu(z)-1/2$ realizes the raw affine separator with one ordinary unit. Both labels occur, so a constant cannot suffice.

Let $D$ be the diagonal matrix with entries $D_v\in\{2,3,4\}$. The matrix $S=D^{1/2}PD^{-1/2}$ is symmetric. Apply Gershgorin's theorem to $P$: row $v$ has disk center $P_{vv}=1/D_v$ and radius $1-1/D_v$. Since $S$ is symmetric and similar to $P$, both matrices have the same real spectrum, contained in $[2\min_v P_{vv}-1,1]\subseteq[-1/2,1]$. Hence $T=(I+S)/2$ has spectrum in $[1/4,1]$. Since $M=D^{-1/2}TD^{1/2}$,
\begin{equation}\label{eq:conditionproof}
 \|M\|_2\le\sqrt2,\qquad
 \|M^{-1}\|_2\le4\sqrt2,\qquad \kappa_2(M)\le8.
\end{equation}

\subsection{Aggregate order and transition lower bounds}
At leaf $c_i$, $M$ weights the leaf and helper by $3/4$ and $1/4$, giving $G_{c_i}=\xi_i=1/2+t_i$. For $2\le i\le N-1$, linear spacing of $t$ and alternation of $\sigma$ give
\begin{equation}\label{eq:helperoutputs}
\begin{aligned}
 G_{h_i}&=\tfrac12-\tfrac1{10}\sigma_i+\tfrac72t_i,\\
 G_{h_1}&=\tfrac13+\frac{2}{25(N+1)},\qquad
 G_{h_N}=\tfrac23-\frac{2}{25(N+1)}.
\end{aligned}
\end{equation}
The leaves lie in $(0.49,0.51)$. Negative internal helpers lie in $(0.365,0.435)$ and positive internal helpers in $(0.565,0.635)$. Their respective endpoint helpers lie outside these intervals. Each same-label internal-helper group is strictly increasing. Thus all rows are distinct, and the sorted labels have exactly $N+1$ transitions.

A score in $\mathcal{H}_m$ is continuous piecewise affine with at most $m$ knots. On any finite ordered sample it has at most $m+1$ prediction transitions. To include zero scores, make a sufficiently small common bias shift toward the fixed zero-tie label. This preserves all finite-sample predictions and makes sample scores nonzero. The shift can also avoid identically zero affine pieces. Each of the at most $m+1$ pieces then contains at most one sign-changing root.

If $e$ predictions are wrong, changing them to the true labels changes at most $2e$ adjacent transition indicators. Consequently
\[
 N+1\le m+1+2e,
 \qquad e\ge\left\lceil (N-m)_+/2\right\rceil.
\]
The inclusion $\mathcal{O}_m\subseteq\mathcal{H}_m$ applies this bound to ordinary heads. Zero error requires $m\ge N$.

\subsection{An ordinary score with exactly $N$ units}
Let $A=2/h=100(N+1)$, and set
\[
 J_1=-3A,\quad J_i=2A(-1)^i\ (2\le i\le N-1),\quad J_N=3A.
\]
In particular, $J_3=-2A$. The ordinary score
\begin{equation}\label{eq:explicit}
 s(z)=9+J_3\relu(\xi_3-z)
       +\sum_{i\ne3}J_i\relu(z-\xi_i)
\end{equation}
uses $N$ units. Its left slope is $2A$, its interior slopes alternate $-A,+A,\ldots,-A$, and its right slope is $2A$. Furthermore,
\[
 s(\xi_1)=9-2A(\xi_3-\xi_1)=9-8=1.
\]
The leaf scores are therefore $s(\xi_i)=\sigma_i$. The left helper block is sufficiently far left to have negative scores; the right block has positive scores. This proves exact ordinary width $N$, including $N=4$.

\subsection{Attaining the entire $\mathcal{H}_m$ curve}
Let $q=\min\{\lfloor m/2\rfloor,N/2\}$. For $q=0$, use $s(z)=A(z-0.45)$; the threshold $0.45$ lies between the negative helper block and the leaves. For $q\ge1$, interpolate scores $\sigma_i$ at the first $2q$ leaf positions, use those positions as the $2q$ knots, and give both exterior tails slope $A$. These interpolants classify all helpers correctly. For $i>2q$ their leaf scores are
\[
 -1+2(i-2q)>0.
\]
Only the $N/2-q$ remaining negative leaves are wrong, exactly $\lceil(N-m)_+/2\rceil$. At most $m$ knots are used, including when $m$ is odd. This matches the lower bound and completes the proof.

\section{Proof of Proposition 1}\label{app:stability}
Among distinct raw feature values the minimum gap is $8h$; raw leaves repeat the same-label values $0.4$ and $0.6$. All aggregate values are distinct, and their minimum gap is exactly $h$. Consecutive sorted gaps are $h$ between leaves, $7h$ within either same-label internal-helper group, $19/600+(13/2)h$ between an endpoint helper and its nearest internal helper, and $11/200+8h$ between a helper block and the leaf block. Missing internal-helper gaps at $N=4$ cause no exception.

Since $M$ is row-stochastic, $\|ME\|_\infty\le\|E\|_\infty\le h/4$. Every output gap remains at least $h/2$, preserving the entire order and its transition lower bounds. Raw scores retain signed margin at least $1/10-\eta_N$, and features remain in $[0,1]$.

The partial interpolants in Appendix~\ref{app:scalar} have Lipschitz constant $A$ and absolute score at least 1 at every original sample. Under the perturbation, their score changes are at most $A\eta_N=1/2$. Every prediction, including an intentionally incorrect one, is retained. The matching lower and upper bounds prove the unchanged exact $\mathcal{H}_m$ curve.

To attain ordinary width $N$ after perturbation, write $x_i=(M\widetilde F)_{c_i}$ and $\Delta_i=x_{i+1}-x_i\in[h/2,3h/2]$. Let $\ell_i=(\sigma_{i+1}-\sigma_i)/\Delta_i$ for $1\le i<N$, set $a=\ell_2-\ell_3=2/\Delta_2+2/\Delta_3\ge8/(3h)$, and put $\ell_0=\ell_N=a$ and $J_i=\ell_i-\ell_{i-1}$. Then $J_3=-a$. The ordinary $N$-unit score
\begin{equation}\label{eq:perturbedscore}
\begin{aligned}
 \widetilde s(z)&=b+J_3\relu(x_3-z)
       +\sum_{i\ne3}J_i\relu(z-x_i),\\
 b&=1+a(x_3-x_1).
\end{aligned}
\end{equation}
has slopes $\ell_i$ between successive leaves, exterior slopes $a$, and $\widetilde s(x_i)=\sigma_i$. The original helper-to-leaf block gap is $11/200+8h$; after perturbation it is at least $11/200+(15/2)h>h$. Left helpers therefore have scores below $1-ah<0$, and right helpers have scores above $-1+ah>0$. This proves the matching ordinary upper bound throughout the stated neighborhood.

\section{Proofs for the Full-Rank Extension}\label{app:highdim}
\subsection{Orthogonality and centered conditioning}
There are $N-1-r$ Walsh characters with at least two bits, allowing $d\le N-r$. Their orthogonality to each other and to every singleton character gives
\begin{equation}\label{eq:walshproducts}
 \langle X_J,X_L\rangle=\tfrac N{10}\mathbf{1}\{J=L\},
 \qquad \langle X_0,X_J\rangle=0.
\end{equation}
Direct sums yield
\[
 \|t\|^2=\frac{N(N-1)}{30000(N+1)},\qquad
 \sigma^{\mathsf T}t=-\frac{N}{100(N+1)}.
\]
Thus
\begin{equation}\label{eq:primarynorm}
 \frac{\|X_0\|^2}{N}
 =\frac1{10}+\frac{16N+704}{30000(N+1)}<0.11.
\end{equation}
All squared singular values of $X$ lie in $[0.1N,0.11N]$, proving~\eqref{eq:xcondition}. All chosen Walsh characters are nonconstant, and $t$ and $\sigma$ sum to zero. Hence every column of $X$ has zero global mean.

For any $x\perp\mathbf{1}_{2N}$, $M\mathbf{1}_{2N}=\mathbf{1}_{2N}$ gives
\begin{align*}
 \|HMx\|_2
 &=\min_{c\in\mathbb{R}}\|Mx-c\mathbf{1}_{2N}\|_2\\
 &=\min_{c\in\mathbb{R}}\|M(x-c\mathbf{1}_{2N})\|_2\\
 &\ge\sigma_{\min}(M)\|x\|_2.
\end{align*}
Also $\|HMx\|_2\le\sigma_{\max}(M)\|x\|_2$. Apply these inequalities to $x=Xb$, $b\in\mathbb{R}^d$, to obtain the full rank and condition bounds for $Y$. For any full-column-rank $B$, $\sr(B)\ge d/\kappa_2(B)^2$, proving the stable-rank bound in~\eqref{eq:ycondition}.

\subsection{The hard leaves and ambient counterexample}
The leaf weights $3/4,1/4$ cancel each nuisance column $(a_J,-3a_J)$, proving~\eqref{eq:leafline}. Restricting an ordinary $d$-input width-$m$ head to this line gives a scalar score with at most $m$ knots. The leaves have $N-1$ true-label transitions, so $N-1\le m+1+2e_c$, where $e_c$ is their error count. This proves~\eqref{eq:multirisk}. The scalar score~\eqref{eq:explicit}, using only the primary coordinate, gives the zero-error upper bound $N$. The primary coordinate also keeps every full row distinct.

For Proposition 3, adding $\epsilon\sigma_i$ to both paired raw entries makes the selected leaf nuisance output $1/2+\epsilon\sigma_i$. The primary coordinate is unchanged. At leaves $z_0\in(0.49,0.51)$, so the last two terms of~\eqref{eq:three} vanish and its score is $\epsilon\sigma_i$. Negative helpers have $z_0<0.435$ and positive helpers $z_0>0.565$. The last two terms then give the required sign with magnitude greater than $0.45$, dominating $|z_1-1/2|\le0.31$. Indeed, the raw nuisance deviation from $1/2$ is at most $3\gamma+\epsilon\le0.31$, and row-stochastic averaging preserves this bound. Since $z_1>0$, the first term equals $z_1$. All rows are correct. Perturbed raw coordinates stay in $[0,1]$. Full column rank is an open condition, and singular values are continuous in the table entries, giving the stated conditioning limit as $\epsilon\downarrow0$.

\section{Proof of the Population Extension}
\label{app:population_overlap}
Put $p_\wedge(z)=\min_{v\in V}p_v(z)$ and
$K_m=\lceil(N-m)_+/2\rceil$. For each $v$, choose a regular conditional law $\mu_{v,z}$ of $A_v$ given $Z_v=z$, defined for $p_v(z)\,dz$-almost every $z$ and extended arbitrarily elsewhere. Almost-sure membership in $I_v$ gives $\mu_{v,z}(I_v)=1$ for $p_v(z)\,dz$-almost every $z$. Since $p_\wedge\le p_v$ and $V$ is finite, all these interval conditions hold together for $p_\wedge(z)\,dz$-almost every $z$. For such a fixed $z$, the restriction $a\mapsto s(a,z)$ is an ordinary scalar score of width at most $m$.

The scalar aggregate minimum gap is $h$. Hence every tuple $(a_v)_{v\in V}\in\prod_{v\in V}I_v$ retains the original value order and its $N+1$ label transitions. The transition argument of Theorem~1 gives
\[
 \sum_{v\in V}\mathbf{1}\{\operatorname{pred}(s(a_v,z))\ne y_v\}\ge K_m.
\]
Integrate this inequality against the auxiliary product measure $\bigotimes_{v\in V}\mu_{v,z}$. If
\[
 e_v(z)=\int_{I_v}\mathbf{1}\{\operatorname{pred}(s(a,z))\ne y_v\}
                  \,\mu_{v,z}(da),
\]
then $\sum_v e_v(z)\ge K_m$. The product measure is only a proof device; the actual vertex draws may be dependent. Conditional expectation and Tonelli's theorem now give
\begin{align*}
 \mathcal R(s)
 &=\frac1{2N}\int_{\mathbb R^q}\sum_v p_v(z)e_v(z)\,dz\\
 &\ge\frac1{2N}\int_{\mathbb R^q}p_\wedge(z)\sum_v e_v(z)\,dz
 \ge\frac{\omega K_m}{2N}.
\end{align*}

For the upper bound, let $s_N$ be the explicit clean ordinary $N$-unit score in~\eqref{eq:explicit}. Its leaf scores are $+1,-1,\ldots,+1,-1$. Its interior slopes have magnitude $2/h$ and its exterior slopes are $4/h$, so its global Lipschitz constant is $4/h$. Each helper block is more than $h/2$ from the nearest leaf, giving signed helper scores greater than one. Thus, for every $a\in I_v$,
\[
 y_vs_N(a)\ge y_vs_N(G_v)-|s_N(a)-s_N(G_v)|
 >1-\frac4h\frac h4=0.
\]
Set all nuisance weights to zero. This fixed $N$-unit score has zero risk, while~\eqref{eq:population_width_bound} is positive for $m<N$. For a random fitted head independent of a fresh evaluation draw, condition on the head and apply the same argument.

\paragraph{An iid raw-noise example.}
Let $F\in\mathbb R^{2N}$ and $M$ be the scalar raw table and row-stochastic lazy mean from Theorem~1. Choose $0<\beta<1/2$. For every vertex $w\in V$ and coordinate $j\in\{1,\ldots,q\}$, draw $U_{w,j}$ independently and uniformly on $(-\beta,\beta)$. Also draw $E_w$ independently and uniformly on $(-\eta_N,\eta_N)$, independently of all the $U_{w,j}$. Form the raw coordinates
\[
 C_{w,0}=F_w+E_w,\qquad C_{w,j}=\tfrac12+U_{w,j},
\]
and apply $M$ to each column. The resulting row is
\[
 A_v=G_v+\sum_wM_{vw}E_w,\qquad
 (Z_v)_j=\tfrac12+\sum_wM_{vw}U_{w,j}.
\]
Because $M_{vw}\ge0$ and $\sum_wM_{vw}=1$, $A_v\in I_v$ almost surely. Repeated convolution of interval densities shows that the primary density is positive throughout $I_v$ and each nuisance marginal is positive throughout $(1/2-\beta,1/2+\beta)$. Independence of the raw coordinates gives a full-dimensional joint row density. Different vertices can remain dependent.

\paragraph{Quantitative common overlap.}
The self weight is $a_v=M_{vv}=(1+D_v^{-1})/2\in\{3/4,2/3,5/8\}$. Write one centered nuisance coordinate as $a_vU_{v,j}+R_{v,j}$, where $R_{v,j}=\sum_{w\ne v}M_{vw}U_{w,j}$ is independent of $U_{v,j}$ and supported on $[-(1-a_v)\beta,(1-a_v)\beta]$. Conditioning on $R_{v,j}$ gives its density
\[
 f_v(x)=\frac{\Pr\{R_{v,j}\in[x-a_v\beta,x+a_v\beta]\}}
                   {2a_v\beta}.
\]
For $|x|\le\beta/4$, $|x-R_{v,j}|\le\beta/4+(1-a_v)\beta\le a_v\beta$, since $a_v\ge5/8$. Hence $f_v(x)=1/(2a_v\beta)\ge2/(3\beta)$ on this interval, up to immaterial boundary density values. Independence across nuisance coordinates gives
\[
 p_v(z)\ge\left(\frac{2}{3\beta}\right)^q
 \quad\text{on}\quad
 \mathcal C=(\tfrac12-\tfrac\beta4,\tfrac12+\tfrac\beta4)^q.
\]
Taking the minimum over vertices and integrating yields
\[
 \omega\ge\left(\frac\beta2\right)^q
              \left(\frac{2}{3\beta}\right)^q=3^{-q}.
\]
Corollary~1 proves~\eqref{eq:uniform_noise_risk}. For $q=m=1$, even $N$ gives $\lceil(N-1)/2\rceil=N/2$, hence risk at least $1/12$. The fixed $N$-unit score remains correct on the open primary intervals. The raw separator remains correct because $\eta_N<1/10$. Thus the population zero-error width increases from 1 to $N$ in this continuous-noise example. The bound on $\omega$ need not be tight.

\raggedbottom
\begin{IEEEbiography}[{\includegraphics[width=1in,height=1.25in,keepaspectratio,trim=325bp 94bp 33bp 40bp,clip]{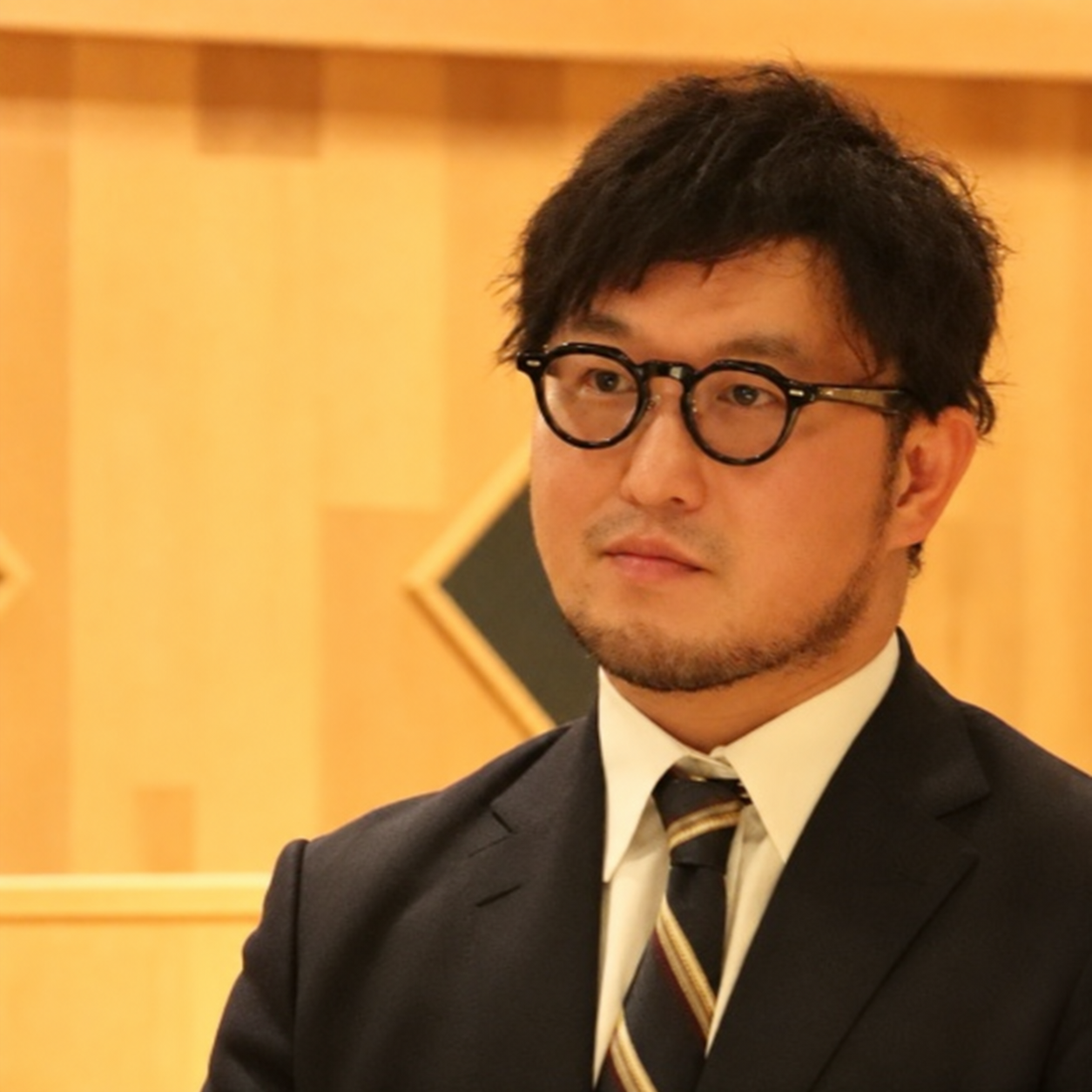}}]{Hiroyasu Usami}
received the Ph.D. degree in engineering from Chubu University in 2019. He is a Senior Assistant Professor at Chubu University and a Researcher at Aichi Medical University. His research focuses on computer vision, machine learning, and medical image analysis. He serves on the Editorial Board of \emph{Scientific Reports} and the editorial committee of the \emph{IPSJ Journal/Journal of Information Processing}.
\end{IEEEbiography}

\begin{IEEEbiography}[{\includegraphics[width=1in,height=1.25in,keepaspectratio,trim=94.909091bp 101.127273bp 194.072727bp 31.090909bp,clip]{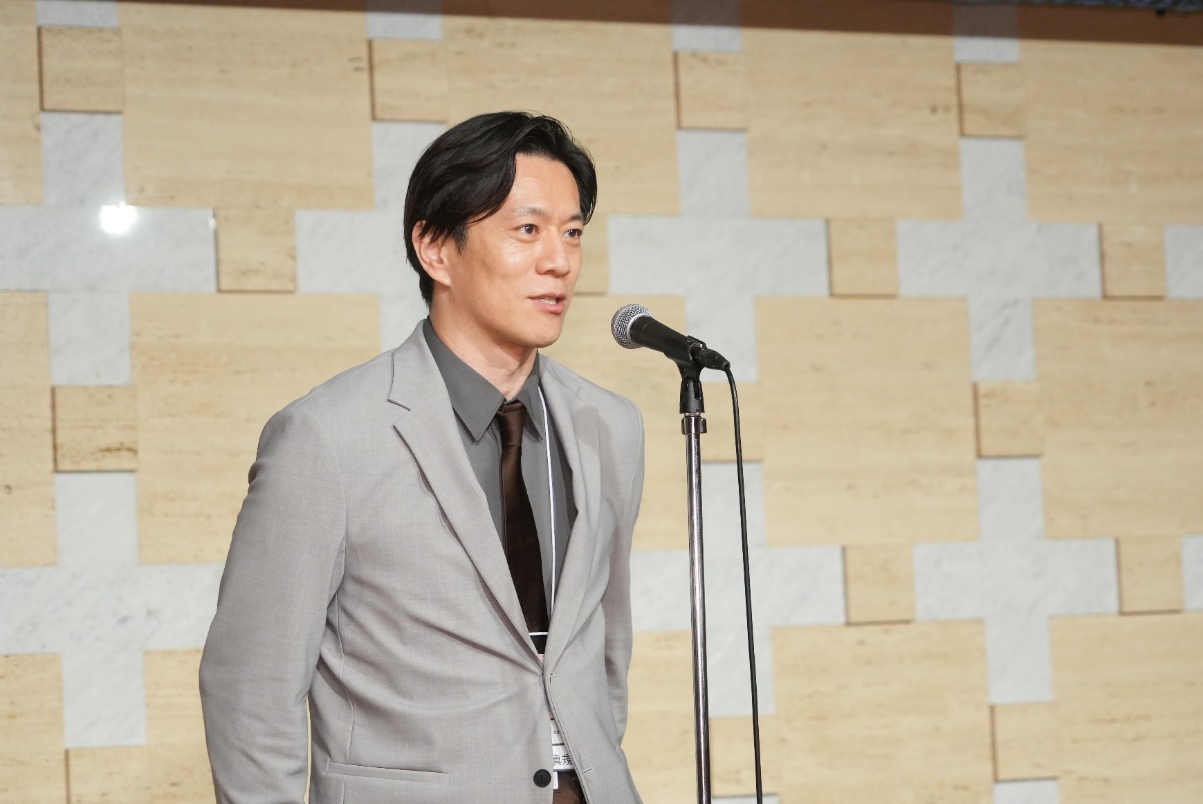}}]{Masanori Tsukamoto}
received the Ph.D. degree from Saitama Medical University, Saitama, Japan, in 2012. He is currently an Associate Professor in the Department of Systemic Management for Dentistry, Kagoshima University Hospital, Kagoshima, Japan. His clinical and research interests focus on clinical anesthesiology.
\end{IEEEbiography}

\begin{IEEEbiography}[{\includegraphics[width=1in,height=1.25in,keepaspectratio,trim=55bp 155bp 55bp 12bp,clip]{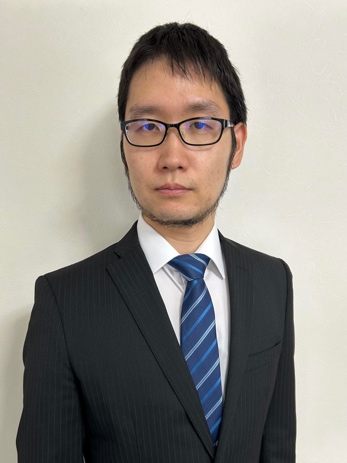}}]{Kohei Hashiguchi}
received the bachelor's degree in dentistry from Kagoshima University in 2017. He is an Assistant Professor at Kagoshima University. His specialty is dental anesthesia, and his research focuses on endotracheal intubation and the airway.
\end{IEEEbiography}
\EOD
\end{document}